\documentclass[final]{cvpr}

\usepackage{amsmath}
\usepackage{color}
\usepackage{colortbl}
\usepackage[pagebackref=true,breaklinks=true,colorlinks,bookmarks=false]{hyperref}
\usepackage{amssymb}
\usepackage{longtable}
\usepackage{xcolor}
\usepackage{siunitx}
\usepackage{pbox}
\usepackage{multirow}
\usepackage{times}
\usepackage{epsfig}
\usepackage{graphbox,graphicx}
\usepackage{subfigure}
\usepackage{verbatim}
\usepackage{booktabs}
\usepackage{diagbox}

\usepackage[pagebackref=true,breaklinks=true,colorlinks,bookmarks=false]{hyperref}

\def\cvprPaperID{5160} %
\def\confYear{CVPR 2021}

\begin{document}

\title{Bottom-Up Human Pose
Estimation Via Disentangled Keypoint Regression}

\author{Zigang Geng$^{1,3}\footnotemark[1]$ , Ke Sun$^{1}\thanks{This work was done when Zigang Geng and
Ke Sun were interns at Microsoft Research, Beijing, P.R. China}$ , Bin Xiao$^{3}$, Zhaoxiang Zhang$^{2}$, Jingdong Wang$^{3}\thanks{Corresponding author}$\\
$^{1}$University of Science and Technology of China\\ $^{2}$Institute of Automation, CAS, University of Chinese Academy of Sciences \\Centre for Artificial Intelligence and Robotics, HKISI\_CAS \\
$^{3}$Microsoft\\
{\tt\small \{zigang,sunk\}@mail.ustc.edu.cn, zhaoxiang.zhang@ia.ac.cn, \{bixi,jingdw\}@microsoft.com}
}

\maketitle
\pagestyle{empty}  
\thispagestyle{empty}

\begin{abstract}
In this paper, we are interested
in the bottom-up paradigm
of estimating human poses from an image.
We study
the dense keypoint regression framework
that is previously inferior
to the keypoint detection and grouping framework.
Our motivation is that
regressing keypoint positions accurately
needs to 
learn representations that focus on the keypoint regions.

We present a simple yet effective approach, named disentangled keypoint regression (DEKR).
We adopt adaptive convolutions 
through pixel-wise spatial transformer
to activate the pixels in the keypoint regions
and accordingly learn representations
from them.
We use a multi-branch structure
for separate regression:
each branch learns a representation 
with dedicated adaptive convolutions
and regresses one keypoint.
The resulting disentangled representations are able to attend
to the keypoint regions,
respectively,
and thus the keypoint regression is spatially more accurate.
We empirically show that the proposed direct regression method outperforms keypoint detection and grouping methods
and achieves superior 
bottom-up pose estimation results on two benchmark datasets, COCO and CrowdPose. The code and models are available at \url{https://github.com/HRNet/DEKR}.

\end{abstract}

\section{Introduction}
Human pose estimation
is a problem of predicting the keypoint positions
of each person from an image,
i.e., localize the keypoints 
as well as identify the keypoints
belonging to the same person.
There are broad applications, 
including action recognition, 
human-computer interaction, smart photo editing,
pedestrian tracking, etc.

There are two main paradigms:
top-down and bottom-up.
The top-down paradigm first detects the person
and then performs single-person pose estimation
for each detected person.
The bottom-up paradigm 
either directly regresses the keypoint positions belonging to the same person,
or detects and groups the keypoints,
such as affinity linking~\cite{CaoSWS17,KreissBA19},
associative embedding~\cite{NewellHD17},
HGG~\cite{JinLXWQOL20}
and HigherHRNet~\cite{cheng2019bottom}.
The top-down paradigm is more accurate
but more costly due to an extra person detection process,
and the bottom-up paradigm, the interest of this paper, is more efficient.

\begin{figure}
	\footnotesize
	\centering
\includegraphics[width=0.49\linewidth, ]{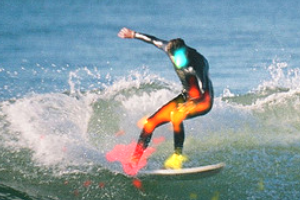}~
\includegraphics[width=0.49\linewidth, ]{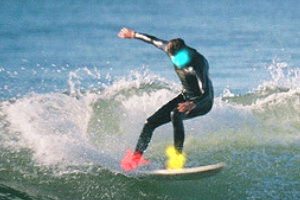}\\
\vspace{.1cm}
\includegraphics[width=0.49\linewidth, ]{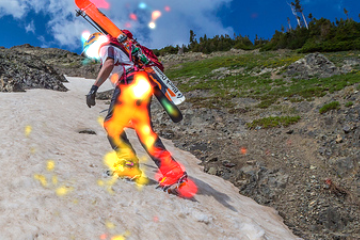}~
\includegraphics[width=0.49\linewidth, ]{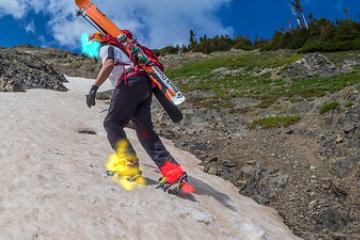}
	\caption{
	Illustration of
	the salient regions
	for regressing the keypoints.
	We take three keypoints, nose and two ankles, as an example for illustration clarity.
	Left: baseline.
	Right: our approach DEKR. 
	It can be seen that 
	our approach is able to focus on the keypoint regions.
	The salient regions are generated using the tool~\cite{uozbulak_pytorch_vis_2019}.}
	\label{fig:teaser}
	\vspace{-0.3cm}
\end{figure}

 \begin{figure*}[t]
 \footnotesize
    \centering
    \includegraphics[height = 0.14\textwidth]{fig/coco_example/id_274460.pdf} 
    \includegraphics[height = 0.14\textwidth]{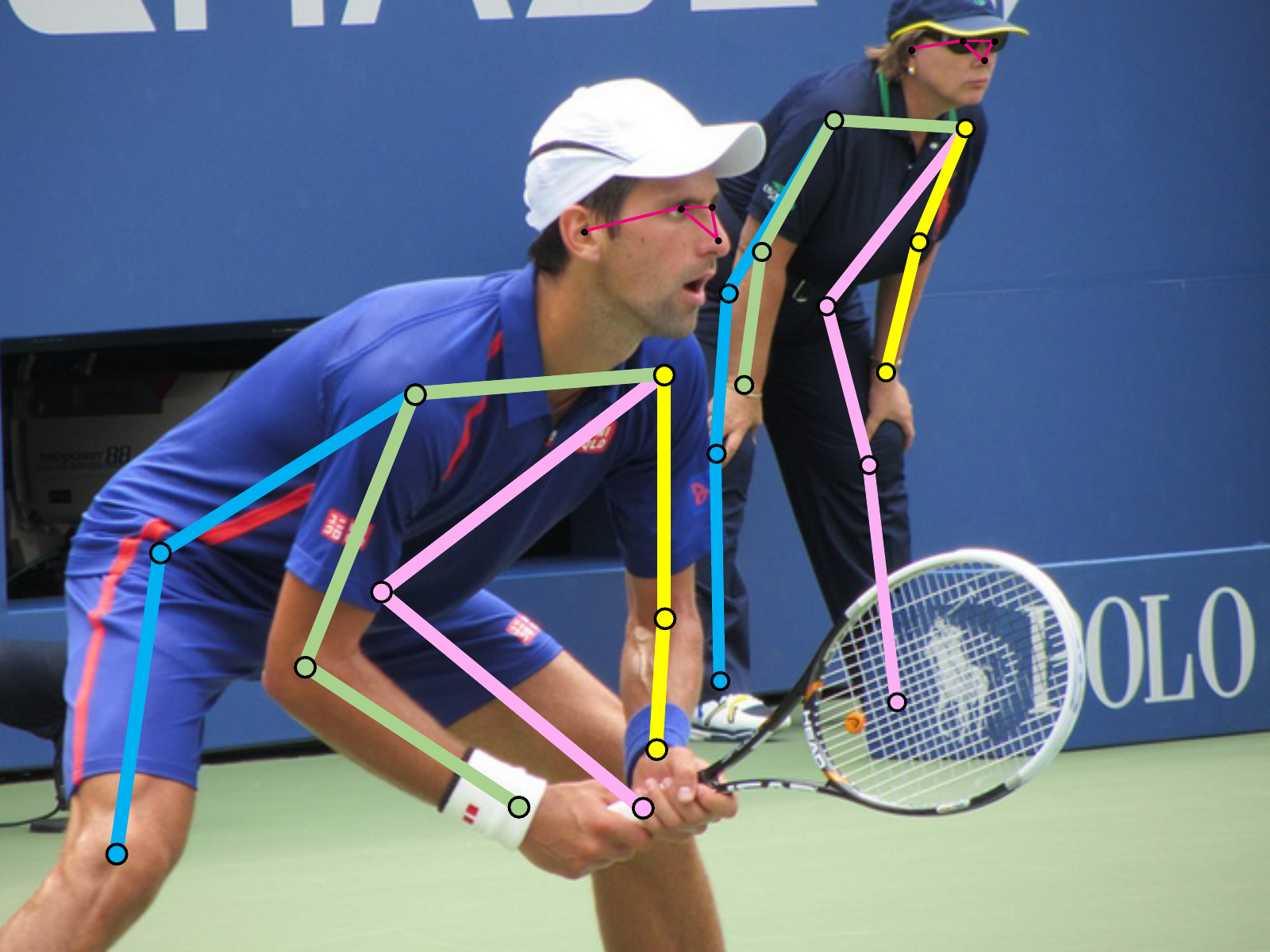}
     \includegraphics[height = 0.14\textwidth]{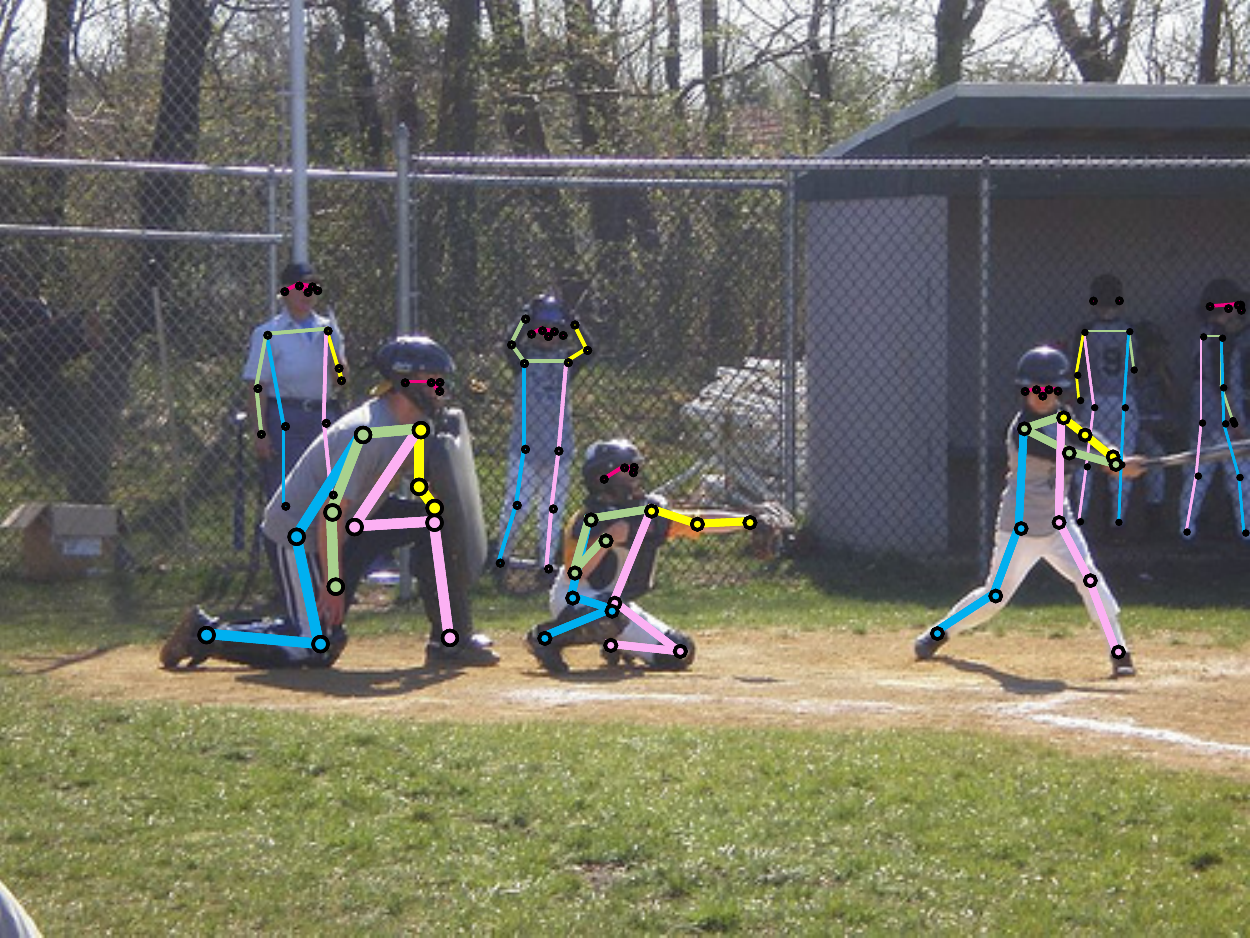}
    \includegraphics[height = 0.14\textwidth]{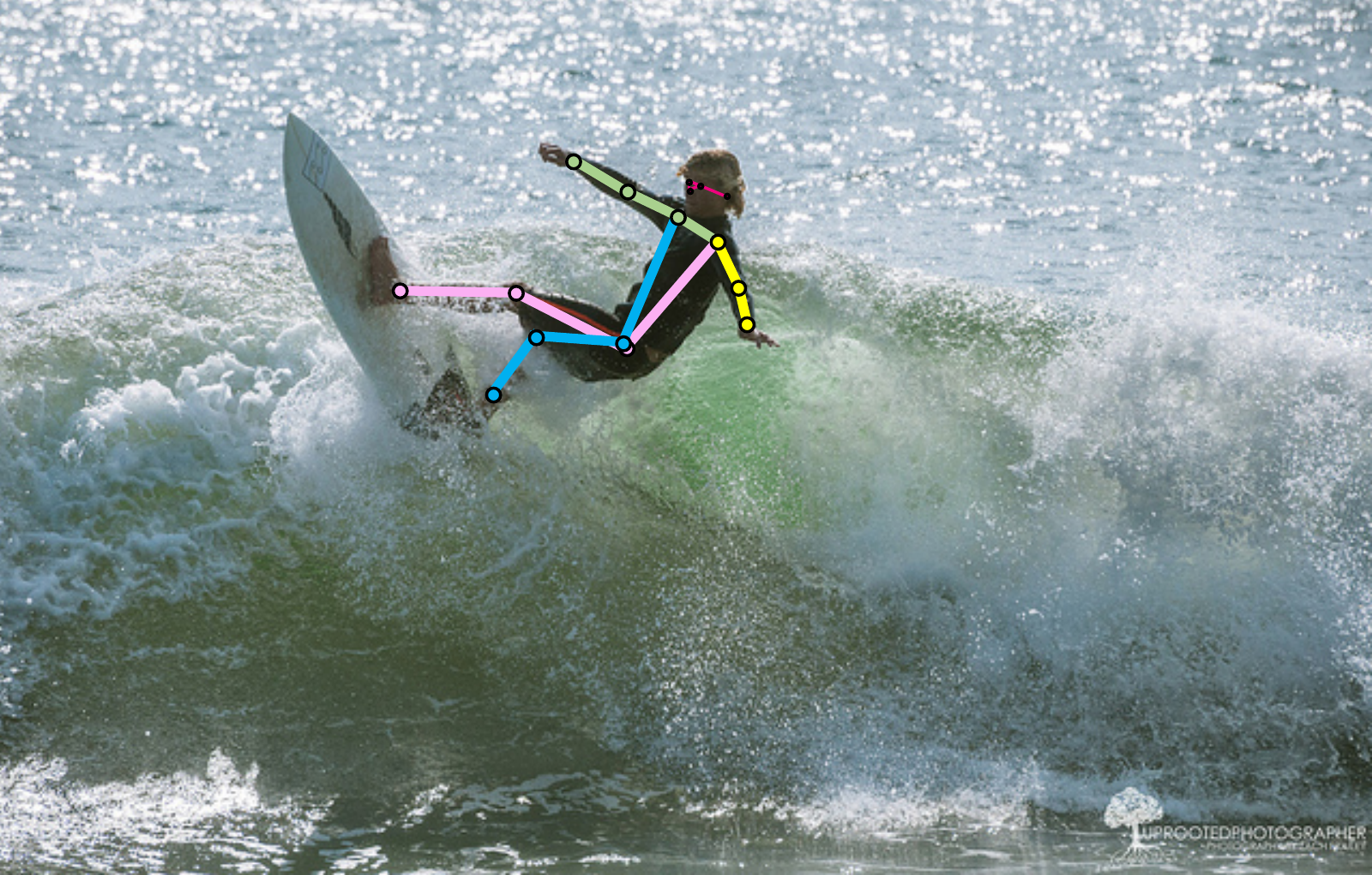}
    \includegraphics[height = 0.14\textwidth]{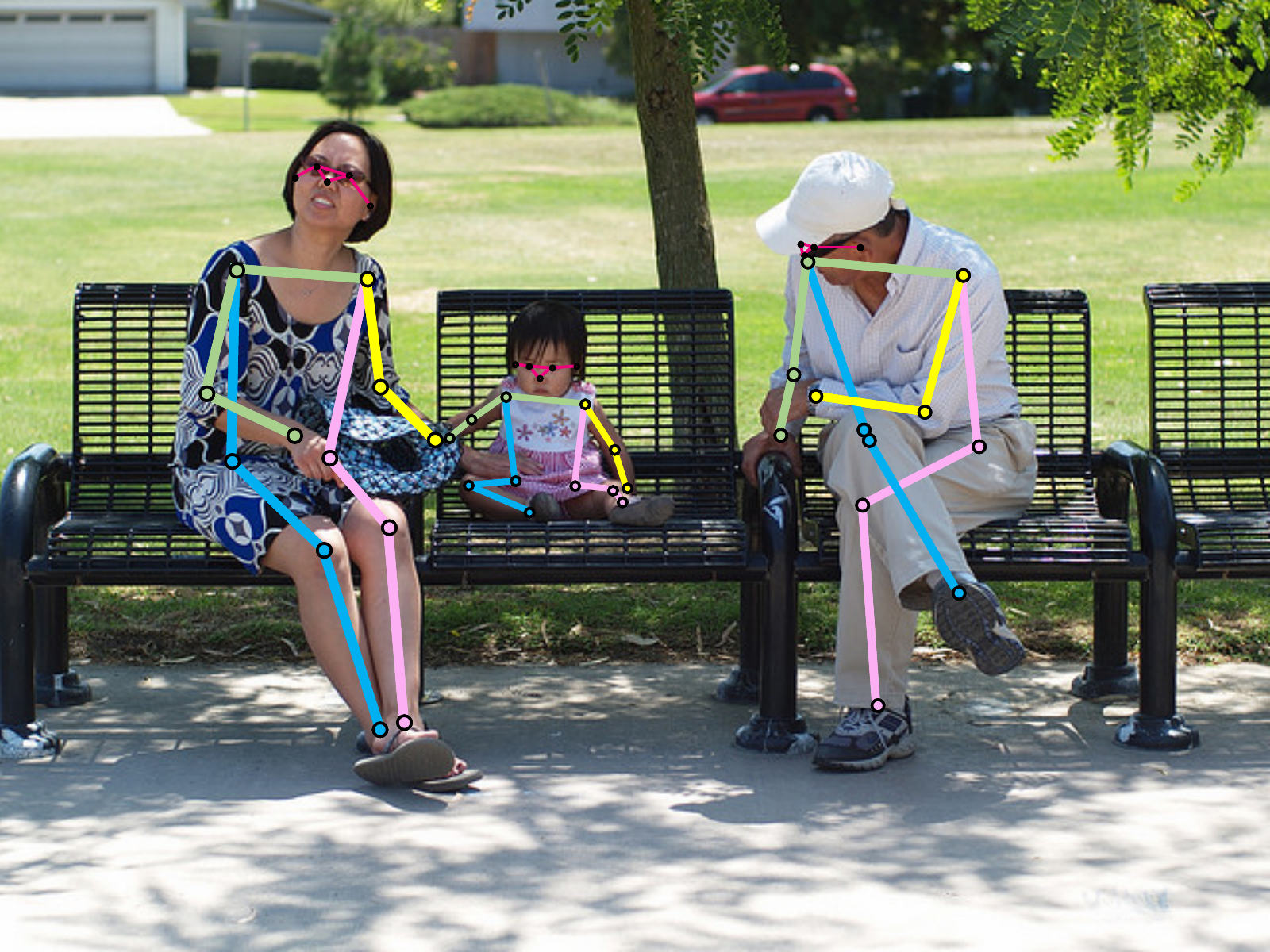}  
    \caption{Multi-person pose estimation. 
    The challenges include
		diverse person scales
		and orientations,
		various poses,
		etc.
		Example results
    are from our approach DEKR.
    }
    \label{fig:humanposeestimationproblem}
    \vspace{-0.3cm}
\end{figure*}

The recently-developed pixel-wise keypoint regression approach, CenterNet~\cite{ZhouWK19}, 
estimates the $K$ keypoint positions
together
for each pixel
from the representation at the pixel.
Direct regression to keypoint positions
in CenterNet~\cite{ZhouWK19}
performs reasonably. 
But the regressed keypoints are spatially not accurate
and the performance is
worse than the keypoint detection and grouping scheme.
Figure~\ref{fig:teaser} (left) shows
two examples in which
the salient areas for keypoint regression
spread broadly
and the regression quality is not satisfactory.

We argue that
regressing the keypoint positions accurately
needs to learn representations that focus on the keypoint regions.
Starting from this \emph{regression by focusing} concept,
we present a simple yet effective approach,
named disentangled keypoint regression (DEKR).
We adopt adaptive convolutions,
through pixel-wise spatial transformer
(a pixel-wise extension of spatial transformer network~\cite{JaderbergSZK15}),
to activate the pixels lying
in the keypoint regions,
and then learn the representations
from these activated pixels,
so that the learned representations
can focus on the keypoint regions.

We further decouple the representation learning
for one keypoint from other keypoints.
We adopt a separate regression scheme
through a multi-branch structure:
each branch learns a representation 
for one keypoint
with adaptive convolutions dedicated
for the keypoint
and regresses the position for the corresponding keypoint. 
Figure~\ref{fig:teaser} (right)
illustrates that
our approach is
able to
learn highly concentrative representations,
each of which focuses on the corresponding keypoint region.

Experimental results demonstrate that
the proposed DEKR approach improves the localization quality of the regressed keypoint positions.
Our approach,
that performs direct keypoint regression
without matching the regression results
to the closest keypoints detected from the keypoint heatmaps,
outperforms keypoint detection and grouping methods
and 
achieves superior performance over previous state-of-the-art bottom-up pose estimation methods
on two benchmark datasets, COCO and CrowdPose.

Our contributions to bottom-up human pose estimation are summarized as follows. 
\begin{itemize}
    \item We argue that
    the representations for regressing the positions
    of the keypoints accurately need to 
    focus on the keypoint regions.
    
    \item 
    The proposed DEKR approach
    is able to learn disentangled representations through two simple schemes, adaptive convolutions
    and multi-branch structure,
    so that each representation focuses on one keypoint region
    and the prediction of the corresponding keypoint position
    from such representation 
    is accurate.
    
    \item The proposed direct regression approach outperforms
    keypoint detection and grouping schemes
    and achieves new state-of-the-art 
    bottom-up pose estimation results
    on the benchmark datasets, COCO and CrowdPose.
\end{itemize}

\section{Related Work}
The convolutional neural network (CNN) solutions~\cite{FanZLW15,LifshitzFU16,SunLXZLW17,YangLOLW17,NieFY18,NieFZY18,PengTYFM18,ZhangZY019,Sekii18}
	to human pose estimation
	have shown superior performance
	over the conventional methods,
	such as the probabilistic graphical model
	or the pictorial structure model~\cite{YangR11,PishchulinAGS13}.
	Early CNN approaches~\cite{ToshevS14,BelagiannisRCN15,CarreiraAFM16} directly predict the keypoint positions for single-person pose estimation,
	which is later surpassed
	by the heatmap estimation based methods~\cite{BulatT16,GkioxariTJ16,ChuYOMYW17,LiuCLQCH18,ArtachoS20}. 
	The geometric constraints and structured relations among body keypoints
	are studied for performance improvement~\cite{ChuOLW16,YangOLW16,ChenSWLY17,TangYW18,KeCQL18,ZhangOLQSYJ19}.

	\vspace{.1cm}
	\noindent\textbf{Top-down paradigm.}
	The top-down methods perform single-person pose estimation
	by firstly detecting each person
	from the image.
	Representative works include:
	HRNet~\cite{SunXLW19, Wang2020},
	PoseNet~\cite{PapandreouZKTTB17}, 
	RMPE~\cite{FangXTL17}, 
	convolutional pose machine~\cite{WeiRKS16},
	Hourglass~\cite{NewellYD16},
	Mask R-CNN~\cite{HeGDG17}, 
	CFN~\cite{HuangGT17}, 
	Integral pose regression~\cite{SunXWLW18}, CPN~\cite{ChenWPZYS18},
	simple baseline~\cite{XiaoWW18},
	CSM-SCARB~\cite{SuYXGW19},
	Graph-PCNN~\cite{WangLGDW20},
	RSN~\cite{CaiWLYDWZZZS20},
	and so on.
	These methods exploit the
	advances in person detection 
	as well as extra person bounding-box labeling information.
	The top-down paradigm, though achieving satisfactory performance, takes extra cost 
	in person box detection.
	
	Other developments
	include improving the keypoint localization 
	from the heatmap~\cite{HuangZGH20, ZhangZDYZ20},
	refining pose estimation~\cite{FieraruKPS18, MoonCL19},
	better data augmentation~\cite{BinCCGTWLHGS20},
	developing a multi-task learning architecture
	combining detection, segmentation and pose estimation~\cite{KocabasKA18},
	and handling the occlusion issue~\cite{li2018crowdpose, QiuZLLWXHC20, ZhouCGWL20}.
	
	\vspace{.1cm}
	\noindent\textbf{Bottom-up paradigm.}
	Most existing bottom-up methods mainly focus on 
	how to associate the detected keypoints 
	that belong to the same person
	together.
	The pioneering work,
	DeepCut~\cite{PishchulinITAAG16}, DeeperCut~\cite{InsafutdinovPAA16},
	and L-JPA~\cite{IqbalG16}
	formulate the keypoint association problem
	as an integer linear program,
	which however takes longer processing time
	(e.g., the order of hours).
	
	Various grouping techniques are developed,
	such as
	part-affinity fields in OpenPose~\cite{CaoSWS17}
	and its extension in PifPaf~\cite{KreissBA19},
	associative embedding~\cite{NewellHD17},
	greedy decoding with hough voting in PersonLab~\cite{PapandreouZCGTM18},
	and graph clustering in HGG~\cite{JinLXWQOL20}.

	Several recent works~\cite{ZhouWK19,NieZYF19,NieFXY18, weiSLW20} 
	densely 
	regress a set of pose candidates,
	where each candidate consists of the keypoint positions
	that might be from the same person.
	Unfortunately,
	the regression quality is not high,
	and the localization quality is weak.
	A post-processing scheme,
	matching the regressed keypoint positions
	to the closest keypoints 
	(which is spatially more accurate) detected from the keypoint heatmaps,
	is usually adopted to improve the regression results.

	Our approach aims to improve
	the direct regression results,
	by exploring our \emph{regression by focusing} idea.
	We learn $K$ disentangled representations, 
	each of which is dedicated for one keypoint
	and learns from the adaptively activated pixels,
	so that each representation focuses
	on the corresponding keypoint area.
	As a result,
	the position prediction 
	for one keypoint
	from the corresponding disentangled representation
	is spatially accurate.
	Our approach is superior to
	and differs from~\cite{VarameshATT20}
	that uses the mixture density network for handling uncertainty 
	to improve direct regression results.

	\vspace{.1cm}
	\noindent\textbf{Disentangled representation learning.}
	Disentangled representations~\cite{BengioCV13} 
	have widely been studied 
	in computer vision~\cite{LorenzBMO19, DentonB17,ZhangT0A0WW19, VillegasYHLL17,ZhuMKG20},
	e.g., disentangling the representations
	into content and pose~\cite{DentonB17},
	disentangling motion from content~\cite{VillegasYHLL17},
	disentangling pose and appearance~\cite{ZhangT0A0WW19}.

	Our proposed disentangled regression
	in some sense can be regarded as
	disentangled representation learning:
	learn the representation 
	for each keypoint separately
	from the corresponding keypoint region.
	The idea of representation disentanglement for pose estimation
	is also explored in the top-down approach,
	part-based branching network (PBN)~\cite{TangW19},
	which learns high-quality heatmaps
	by disentangling representations into each part group.
	They are clearly different: our approach
	learns representations focusing on each keypoint region
	for position regression,
	and PBN de-correlates the appearance representations among different part groups.

\section{Approach}
	Given an image $\mathsf{I}$,
	multi-person pose estimation 
	aims to
	predict the human poses,
	where each pose consists of $K$ keypoints,
	such as shoulder, elbow, and so on.
	Figure~\ref{fig:humanposeestimationproblem}
	illustrates the multi-person pose estimation problem.

\subsection{Disentangled Keypoint Regression}
The pixel-wise keypoint regression framework 
estimates a candidate pose at each pixel $\mathbf{q}$ (called center pixel),
by predicting an $2K$-dimensional offset vector $\mathbf{o}_q$ from the center pixel $\mathbf{q}$ for the $K$ keypoints.
The offset maps $\mathsf{O}$,
	containing the offset vectors
	at all the pixels,
	are estimated through a keypoint regression head,
\begin{align}
    \mathsf{O} & = \mathcal{F}(\mathsf{X}),
\end{align}
where $\mathsf{X}$ is the feature computed from a backbone,
HRNet in this paper,
and $\mathcal{F}(~)$
is the keypoint position
regression head 
predicting the offset maps $\mathsf{O}$.

\begin{figure}[t]
\centering
  \includegraphics[width=1\linewidth]{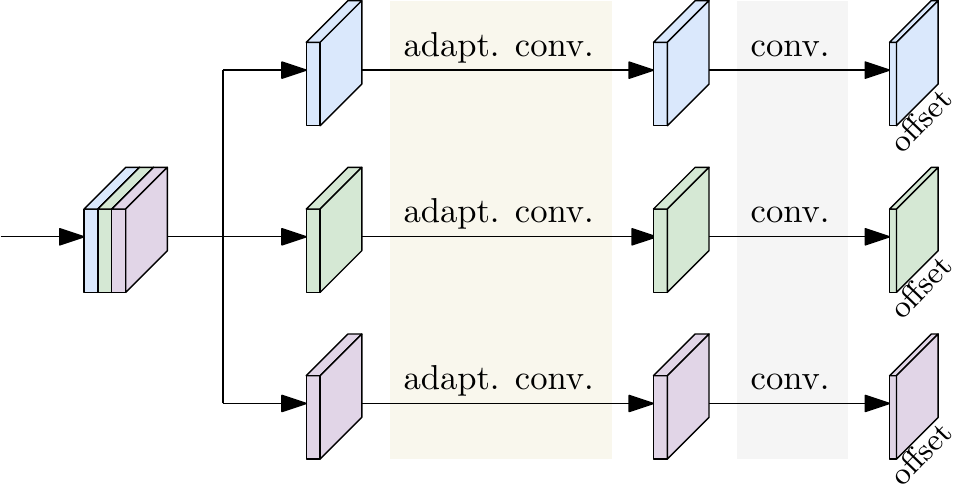}
   \caption{Disentangled keypoint regression.
   Each branch learns the representation for one keypoint
   through two adaptive convolutions
   from a partition of the feature maps
   output from the backbone
   and regresses the $2$D offset of each keypoint using a $1\times 1$ convolution separately.
   This is an illustration for three keypoints,
   and the feature maps are divided into three partitions,
   each fed into one branch.
   In our experiments on COCO pose estimation,
   the feature maps are divided into $17$ partitions 
   and there are $17$ branches for regressing the $17$ keypoints.}
\label{fig:adaptivedisentangledhead}
\vspace{-0.3cm}
\end{figure}

The structure of the proposed 
disentangled keypoint regression (DEKR) head 
is illustrated in Figure~\ref{fig:adaptivedisentangledhead}.
DEKR adopts the multi-branch parallel
adaptive convolutions 
to learn disentangled representations
for the regression of the $K$ keypoints,
so that each representation focuses on
the corresponding keypoint region.

\vspace{.1cm}
\noindent\textbf{Adaptive activation.}
One normal convolution (e.g., $3\times 3$ convolution)
only sees the pixels nearby
the center pixel $\mathbf{q}$.
A sequence of several normal convolutions
may see the pixels farther from the center pixel
that might lie in the keypoint region,
but might not focus on 
and highly activate these pixels.

We adopt the adaptive convolutions,
to learn representations
focusing on the keypoint region.
The adaptive convolution
is a modification of a normal convolution
(e.g., $3 \times 3$ convolution):
	$$\mathbf{y}(\mathbf{q})= \sum\nolimits_{i=1}^9 \mathbf{W}_i \mathbf{x}(\mathbf{g}^q_{si}+ \mathbf{q}). $$
	Here, 
	$\mathbf{q}$ is the center ($2$D) position,
	and $\mathbf{g}^q_{si}$ is the offset,
	$\mathbf{g}^q_{si}+ \mathbf{q}$ corresponds to the $i$th activated pixel.
	$\{\mathbf{W}_1, \mathbf{W}_2, \dots,
	\mathbf{W}_9\}$ are the kernel weights.
	
The offsets $\{\mathbf{g}^q_{s1}, \mathbf{g}^q_{s2}, 
	\dots, \mathbf{g}^q_{s9}\}$
(denoted by a $2\times 9$ matrix $\mathbf{G}_s^q$)
can be estimated by an extra normal $3\times 3$ convolution 
in a nonparametric way like deformable convolutions~\cite{DaiQXLZHW17},
or in a parametric way extending
the spatial transformer network~\cite{JaderbergSZK15}
from a global manner
to a pixel-wise manner.
We adopt the latter one
and estimate an affine transformation matrix $\mathbf{A}^q$ ($\in \mathbb{R}^{2\times 2}$) and a translation vector $\mathbf{t}$ ($\in \mathbb{R}^{2\times 1}$) for each pixel. Then
$\mathbf{G}^q_s = \mathbf{A}^q\mathbf{G}_t +{[\mathbf{t}~\mathbf{t}~\dots~\mathbf{t}]}$.
$\mathbf{G}_t$ represents the regular $3\times 3$ position (meaning that a normal convolution is conducted in the transformed space),
	\makeatletter
	\renewcommand*\env@matrix[1][c]{\hskip -\arraycolsep
		\let\@ifnextchar\new@ifnextchar
		\array{*\c@MaxMatrixCols #1}}
	\makeatother
	\begin{align}
	\mathbf{G}_t = \begin{bmatrix}[r]
	-1 & ~0 & ~1 & -1 & ~~0 & ~~1 & -1 & ~~0 & ~~1 \\[0.3em]
	-1 & -1 & -1 & ~0 & ~~0 & ~~0 & ~1 & ~~1 & ~~1 \nonumber
	\end{bmatrix}.
	\end{align}

\begin{figure*}[t]
\centering
    \footnotesize
    (a)~~\includegraphics[width=0.167\textwidth]{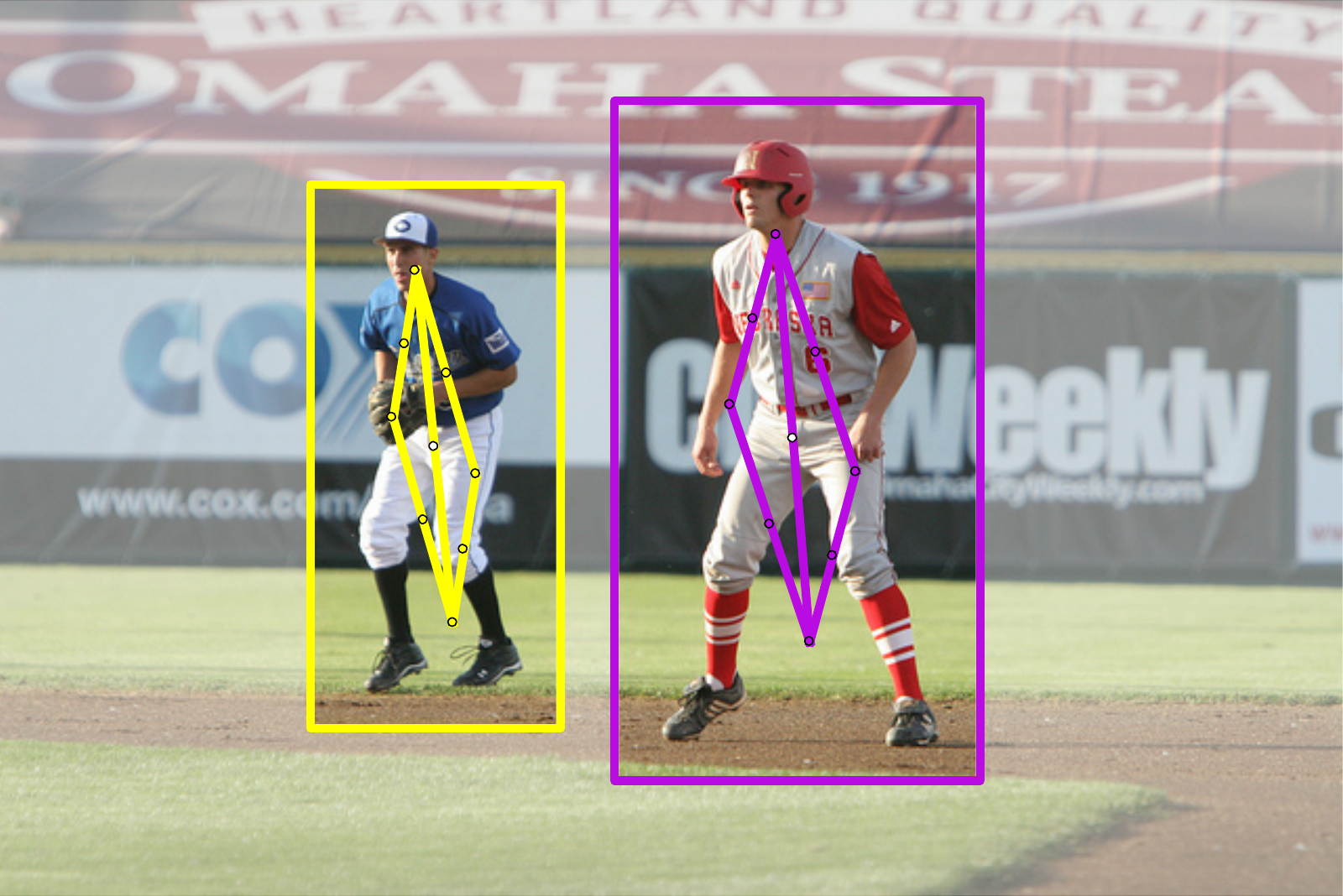}
    ~(b)~~\includegraphics[width=0.167\textwidth]{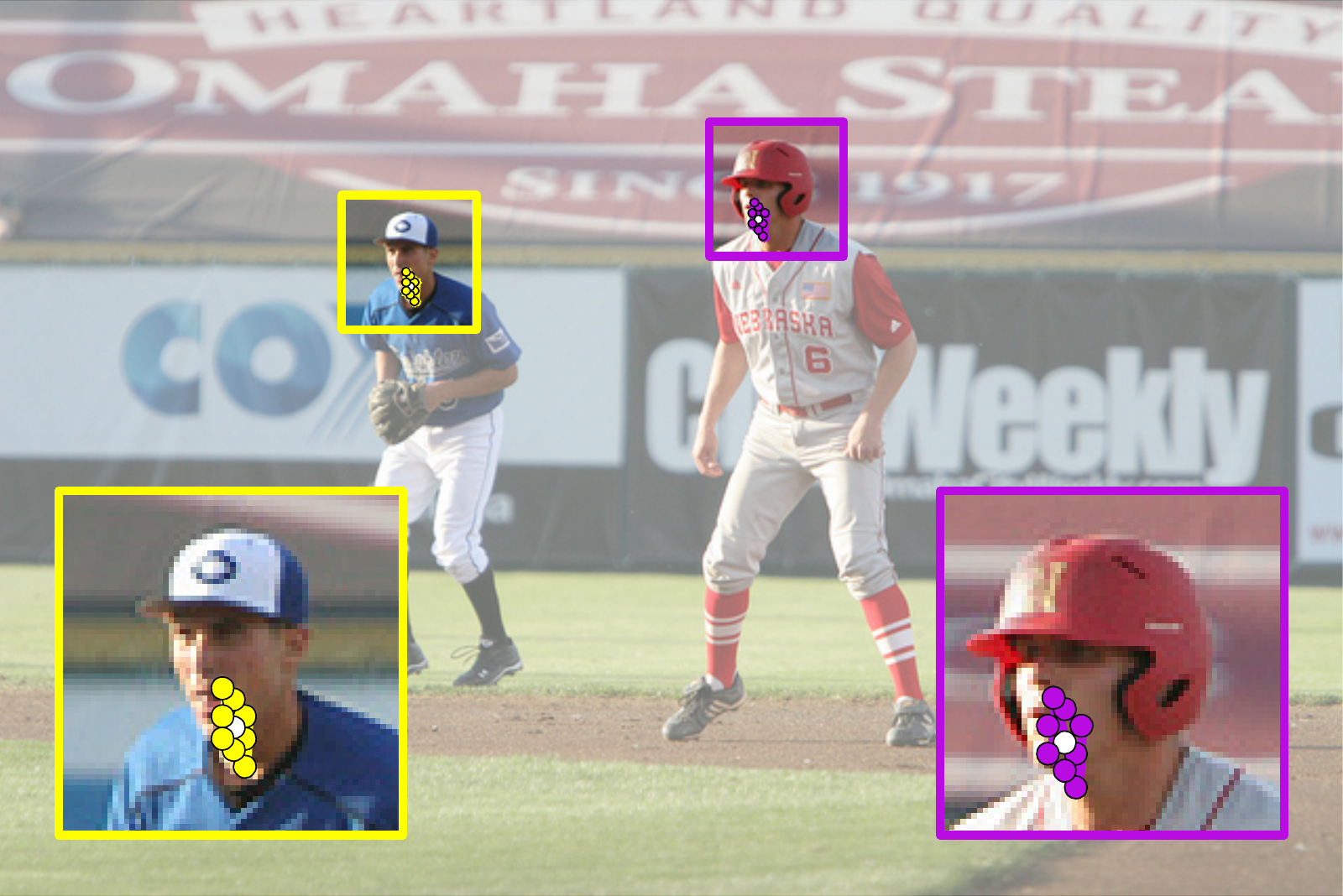}
    ~(c)~~\includegraphics[width=0.167\textwidth]{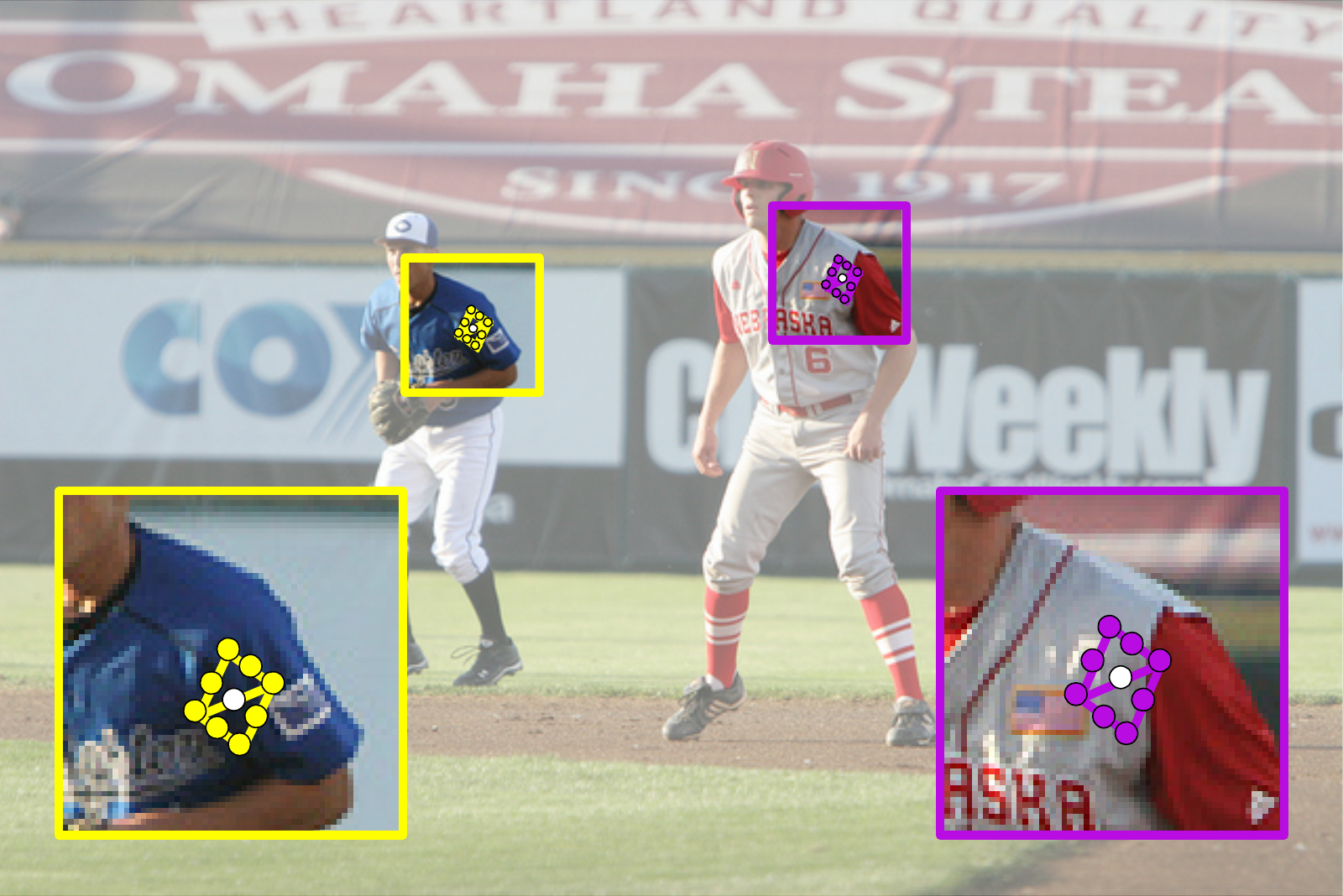}
    ~(d)~~\includegraphics[width=0.167\textwidth]{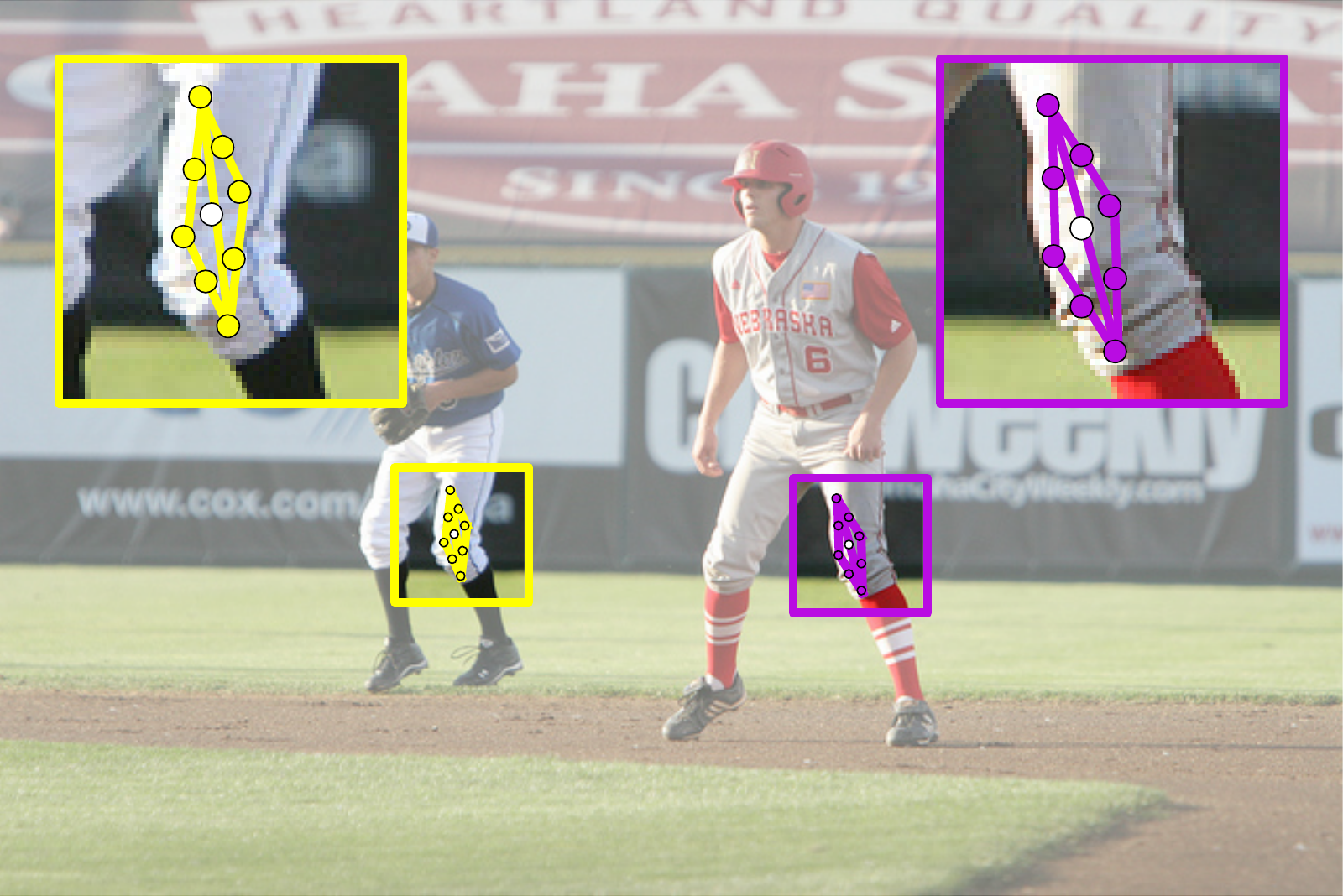}
    ~(e)~~\includegraphics[width=0.167\textwidth]{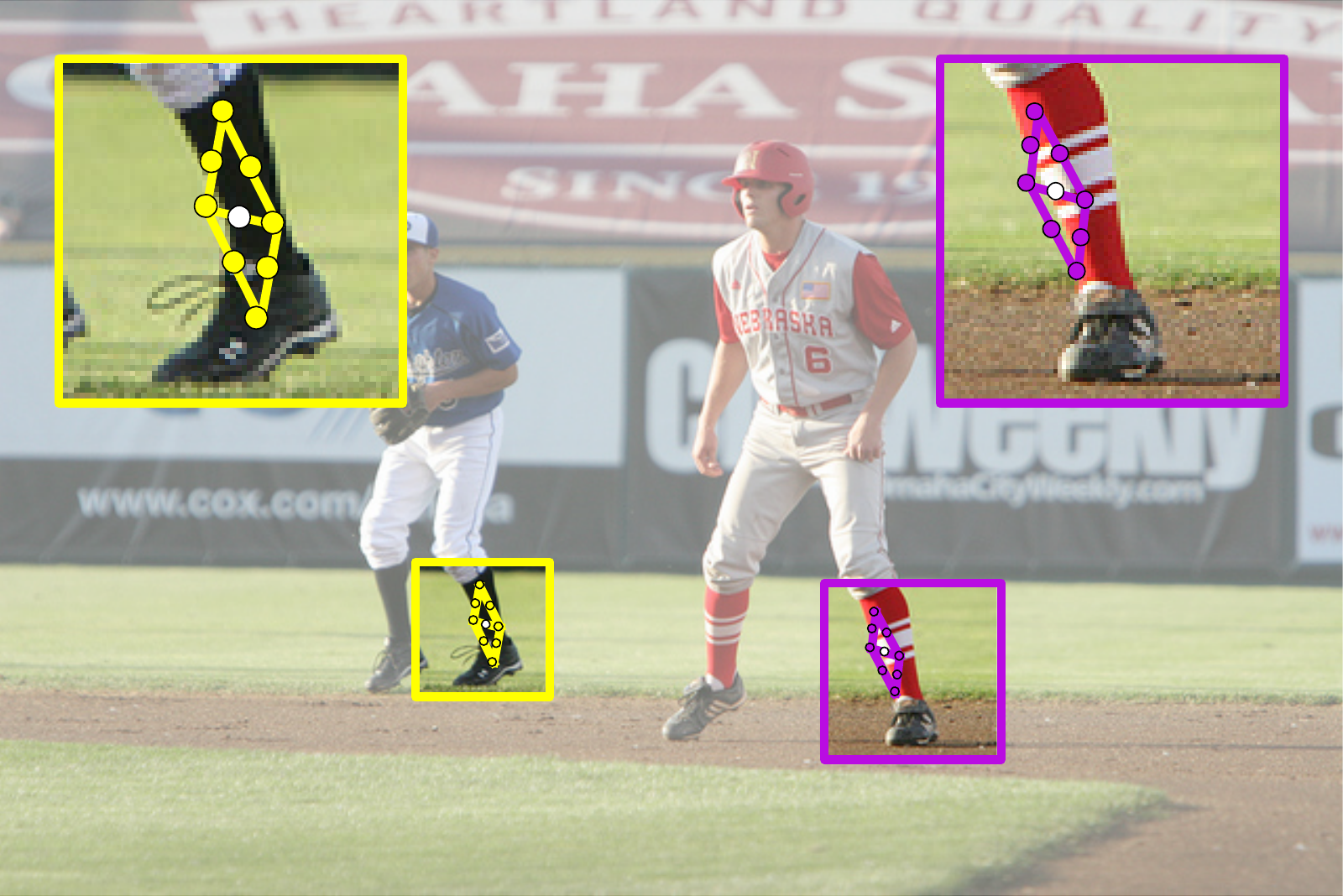} 
   \caption{Illustrating
   adaptive activation.
   (a) Activated pixels from the single-branch regression.
   (b - e) Activated pixels for nose, left shoulder, left knee, and left ankle 
   from the multi-branch regression
   (our approach)
   at the center pixel for each person.
   One can see that the proposed approach
   is able to activate the pixels around the keypoint.
   The illustrations are obtained using the backbone HRNet-W$32$.
   }
\label{fig:adaptiveconvolutionwithmultibranch}
\vspace{-.3cm}
\end{figure*}

\begin{figure}
\centering
\footnotesize
 \includegraphics[width=0.9\linewidth,clip]{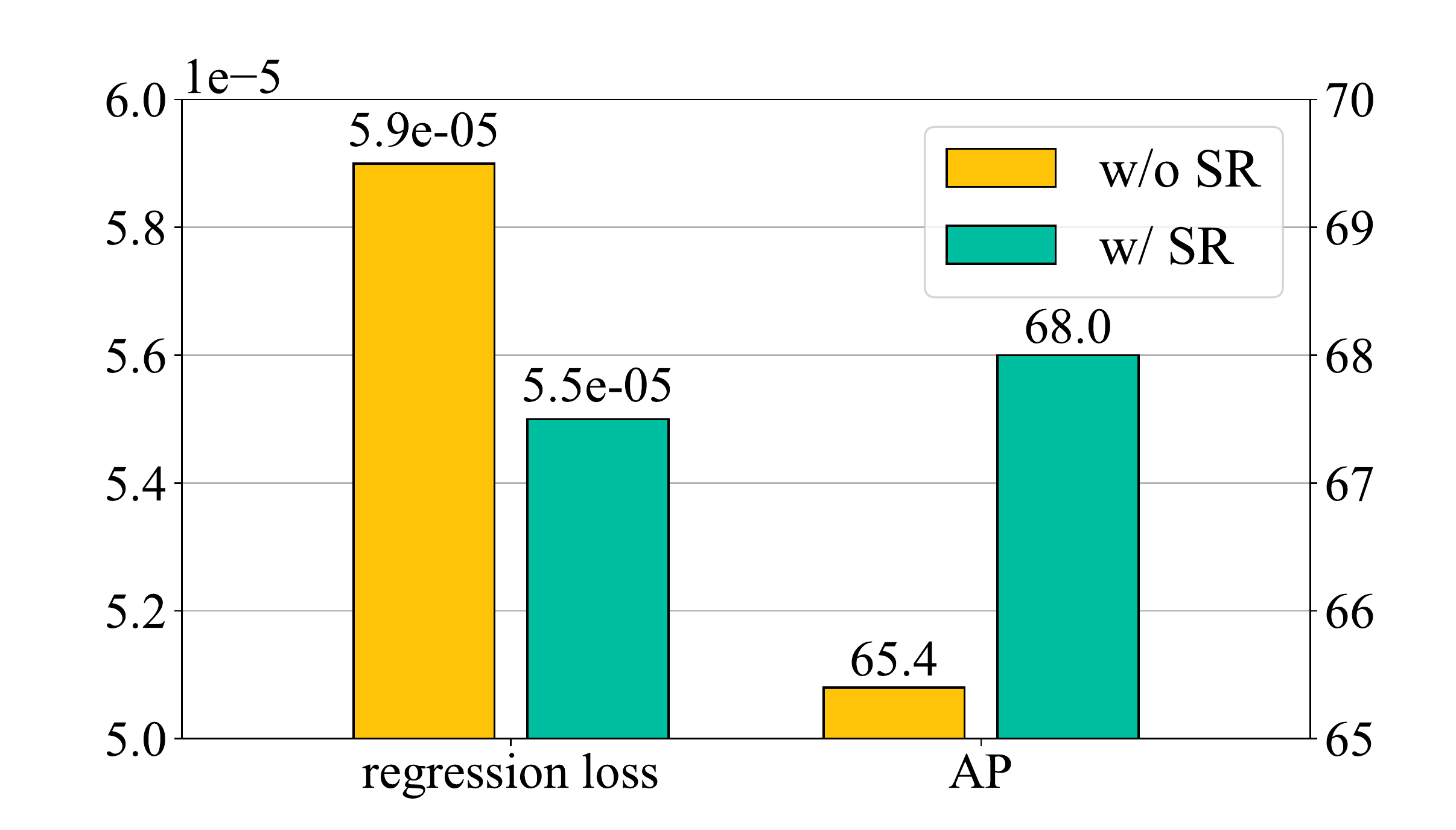} 
   \caption{Separate regression
   improves the regression quality and thus the performance.
   Using separate regression, the regression loss on COCO train set is reduced from $5.9e\text{-}5$ to $5.5e\text{-}5$, and the AP score 
   on COCO validation set is increased from $65.4$ to $68.0$. SR = separate regression. The results are obtained using the backbone HRNet-W$32$.
   }
   \label{fig:disentanglehelpoptimization}
   \vspace{-.3cm}
\end{figure}

\vspace{.1cm}
\noindent\textbf{Separate regression.}
The offset regressor $\mathcal{F}$
in~\cite{ZhouWK19} 
is a single branch,
and estimates
all the $K$ $2D$ offsets
together
from a single feature
for each position.
We propose to use a $K$-branch structure,
where each branch performs the adaptive convolutions
and then regresses the offset for
the corresponding keypoint.

We divide the feature maps $\mathsf{X}$
output from the backbone
into $K$ feature maps,
$\mathsf{X}_1, \mathsf{X}_2,
\dots, \mathsf{X}_K$,
and estimate the offset map $\mathsf{O}_k$ for each keypoint
from the corresponding feature map:
\begin{align}
\mathsf{O}_1 & = \mathcal{F}_1(\mathsf{X}_1) \\
\mathsf{O}_2 & = \mathcal{F}_2(\mathsf{X}_2) \\
& ~~~~~~\vdots \nonumber \\
\mathsf{O}_K & = \mathcal{F}_K(\mathsf{X}_K),
\end{align}
where $\mathcal{F}_k(~)$
is the $k$th regressor
on the $k$th branch,
and $\mathsf{O}_k$
is the offset map for the $k$th keypoint.
The $K$ regressors,
$\{\mathcal{F}_1(~),
\mathcal{F}_2(~),
\dots,
\mathcal{F}_K(~)\}$
have the same structures,
and their parameters are learned independently.

Each branch
in separate regression
is able to learn its own adaptive convolutions,
and accordingly focuses on activating the pixels 
in the corresponding keypoint region 
(see Figure~\ref{fig:adaptiveconvolutionwithmultibranch} {(b - e)}).
In the single-branch case, 
the pixels around all the keypoints are activated,
and the activation is not focused
(see Figure~\ref{fig:adaptiveconvolutionwithmultibranch} (a)).

The multi-branch structure explicitly 
decouples the representation learning
for one keypoint 
from other keypoints,
and thus improves the regression quality.
In contrast, the single-branch structure
has to decouple the feature learning implicitly
which increases the optimization difficulty.
Our results in Figure~\ref{fig:disentanglehelpoptimization}
show that the multi-branch structure
reduces the regression loss.

\subsection{Loss Function}
\noindent\textbf{Regression loss.}
We use the normalized smooth loss
	to form the pixel-wise keypoint regression loss:
	\begin{align}
	\ell_p = \sum\nolimits_{i \in \mathcal{C}}
	\frac{1}{Z_i} \operatorname{smooth}_{L_1}(\mathbf{o}_i - \mathbf{o}^*_i).
	\end{align}
	Here, $Z_i = \sqrt{H_i^2+W_i^2}$ is the size
	of the corresponding person instance,
	and $H_i$ and $W_i$ are the height and the width of 
	the instance box.
	$\mathcal{C}$ is the set
	of the positions that have groundtruth poses.
	$\mathbf{o}_i$ ($\mathbf{o}_i^*$),
	a column of the offset maps $\mathsf{O}$
	($\mathsf{O}^*$),
	is the $2K$-dimensional estimated (groundtruth) offset vector 
	for the position $i$. 
	
\vspace{.1cm}
\noindent\textbf{Keypoint and center heatmap estimation loss.}
We also estimate $K$ keypoint heatmaps
each corresponding to a keypoint type
and the center heatmap indicating 
the confidence that each pixel is the center of
some person,
using a separate heatmap estimation branch,
\begin{align}
(\mathsf{H}, \mathbf{C}) = \mathcal{H}(\mathsf{X}).
\end{align}
The heatmaps are used
for scoring and ranking the regressed poses. 
	The heatmap estimation loss function is formulated 
	as the weighted distances
	between the predicted heat values
	and the groundtruth heat values:
	\begin{align}
	\ell_h = \| \mathsf{M}^h \odot (\mathsf{H} - \mathsf{H}^*)\|_2^2 
	+ \|\mathbf{M}^c \odot(\mathbf{C} - \mathbf{C}^*)\|_2^2.
	\end{align}
	Here, $\|\cdot\|_2$ is the entry-wise $2$-norm.
	$\odot$ is the element-wise product operation.
	$\mathsf{M}^h$ has $K$ masks,
	and the size is $H \times W \times K$.
	The $k$th mask, $\mathsf{M}^h_k$,
	is formed 
	so that the mask weight 
	of the positions not lying in the $k$th keypoint region is $0.1$,
	and others are $1$.
	The same is done for the mask $\mathbf{M}^c$
	for the center heatmap.
	$\mathsf{H}^*$ and $\mathbf{C}^*$
	are the target keypoint and center heatmaps.

\vspace{.1cm}
\noindent\textbf{Whole loss.}
The whole loss function
	is the sum of the heatmap loss
	and the regression loss:
	\begin{align}
	\ell = \ell_h + \lambda \ell_p,
	\end{align}
	where $\lambda$
	is a trade-off weight,
	and set as $0.03$
	in our experiments.

\subsection{Inference}
A testing image is fed into the network,
outputting the regressed pose at each position,
and the keypoint and center heatmaps.
We first perform the center NMS process on the center heatmap
to remove non-locally maximum positions
and the positions whose center heat value is not higher than $0.01$.
Then we perform the pose NMS process 
over the regressed poses at the positions
remaining after center NMS,
to remove
some overlapped regressed poses,
and maintain at most $30$ candidates.
The score used in pose NMS is the average of the heat values at the regressed $K$ keypoints,
which is helpful to keep candidate poses with highly accurately localized keypoints.

We rank the remaining candidate poses
using the score that
is estimated
by jointly considering their corresponding center heat values, keypoint heat values
and their shape scores.
The shape feature
includes
	the distance
	and the relative offset
	between a pair of neighboring keypoints\footnote{
	A neighboring pair $(i, j)$
	corresponds to a stick in the COCO dataset,
	and there are $19$ sticks (denoted by
	$\mathcal{E}$) in the COCO dataset.}:
	$\{d_{ij} | (i, j) \in \mathcal{E}\}$
	and $\{\mathbf{p}_i - \mathbf{p}_j| (i, j) \in \mathcal{E}\}$, and keypoint heat values indicating the visibility of each keypoint.
We input the three kinds of features
to a scoring net, 
consisting of two fully-connected layers
	(each followed by a ReLU layer),
	and a linear prediction layer,
which aims to learn the OKS score
	for the corresponding predicted pose
	with the real OKS
	as the target
	on the training set.

\subsection{Discussions}
\label{sec:approach:dicussions}

\noindent\textbf{Separate regression,
group convolution and complexities.}
In the multi-branch structure, 
we divide the channel maps into $K$ 
non-overlapping partitions,
and feed each partition into each branch,
for learning disentangled representations.
This process resembles group convolution.
The difference lies in: group convolution 
usually increases the capacity 
of the whole representation through 
reducing the redundancy 
and increasing the width
within the computation and parameter budget,
while our approach does not change the width
and aims to learn rich representations focusing on each keypoint.

Let's look at an example with HRNet-W$32$ as the backbone.
The standard process in HRNet-W$32$ concatenates the channels obtained from $4$ resolutions and feeds the concatenated channels
to a $1\times 1$ convolution, outputting $256$ channels. 
When applied to our disentangled regressors,
we modify the $1\times 1$ convolution
to output $255$ ($= 17 \times 15$) channels,
so that each partition has $15$ channels.
This modification does not increase the width.
The parameter complexity
and the computation complexity
for the regression head 
are reduced,
and in particular 
the overall computation complexity
is significantly reduced.
The detailed numbers are given in Table~\ref{tab:disentanglereducecomplexity}.

\vspace{.1cm}
\noindent\textbf{Separate group regression.}
It is noticed that
the salient regions and the activated pixels for some keypoints
might have some overlapping.
For example,
the salient regions of the five keypoints in the head 
are overlapped,
and three keypoints in the arms have similar characteristics.
We investigate the performance if grouping some keypoints
into a single branch
instead of letting each branch handle one single keypoint.
We consider two grouping schemes.
(1) Five keypoints in the head
use a single branch, and there are totally $13$ branches.
(2)
Five keypoints in the head
use a single branch, 
the three keypoints in left arm 
(right arm, left leg, right leg)
use a single branch.
There are totally $5$ branches.
Empirical results show
that separate group regression performs worse
than separate regression,
e.g., the AP score for five branches
decreases by $0.4$ on COCO validation with the backbone HRNet-W$32$.

\renewcommand{\arraystretch}{1.2}
    \begin{table}[t]
	    \centering
	    \setlength{\tabcolsep}{7.02pt}
	    \footnotesize
	    \caption{
	    Comparing the parameter and
	    computation complexities
	    between disentangled keypoint regression
	    (DEKR),
	    baseline regression (baseline),
	    baseline + adaptive activation (+ AA),
	    baseline + separation regression (+ SR).
	    Head: only the regression head is counted.
	    Overall: the whole network is counted.
	    The statistical results are from the backbone HRNet-W$32$.
	    }
	    \label{tab:disentanglereducecomplexity}
	    \begin{tabular}{l|c|c|c|c}
	        \noalign{\smallskip}
	         \hline
	        \multirow{2}{*}{Method}& \multicolumn{2}{c|}{Head} & \multicolumn{2}{c}{Overall} \\
	        \cline{2-5}
	         & \#param. (M) & GFLOPs & \#param. (M) & GFLOPs \\
	        \hline
	        baseline  & $1.31$ & $21.48$ & $30.65$ & $63.28$\\
	        \hline
	        + AA  & $1.34$ & $21.94$ & $30.68$ & $63.73$\\
	        \hline
	        + SR  & $0.19$ & $3.14$ & $29.53$ & $44.93$\\
	        \hline
	        DEKR & $0.22$ & $3.59$ & $29.56$ & $45.39$\\
	        \hline
	    \end{tabular} 
		\vspace{-.3cm}
	\end{table}

\renewcommand{\arraystretch}{1.15}
	\begin{table*}[t]
		\centering
		\caption{Comparisons on the COCO validation set. AE: Associative Embedding~\cite{NewellHD17}.}
\setlength{\tabcolsep}{11.24pt}
		\label{table:coco_val}
		\footnotesize
		\begin{tabular}{l|c|lllll|lll}
		    \hline
			Method & Input size &$\operatorname{AP}$ & $\operatorname{AP}^{50}$ & $\operatorname{AP}^{75}$ & $\operatorname{AP}^{M}$ & $\operatorname{AP}^{L}$ & $\operatorname{AR}$ & $\operatorname{AR}^{M}$ & $\operatorname{AR}^{L}$\\
			\hline
			\multicolumn{10}{c}{single-scale testing}\\
			\hline
			CenterNet-DLA~\cite{ZhouWK19} & $512$ &$58.9$ & $-$ & $-$ & $-$ & $-$&$-$ &$-$ & $-$\\
			CenterNet-HG~\cite{ZhouWK19} & $512$ &$64.0$ & $-$ & $-$ & $-$ & $-$ & $-$ & $-$& $-$\\
			PifPaf~\cite{KreissBA19} & $-$   
			&$67.4$ & $-$ & $-$ & $-$ & $-$ & $-$ & $-$& $-$\\
			HGG~\cite{JinLXWQOL20} & $512$ & $60.4$ & $83.0$ & $66.2$ & $-$ & $-$ & $64.8$ & $-$ & $-$\\
			PersonLab~\cite{PapandreouZCGTM18} & 601 &$54.1$ & $76.4$ & $57.7$ & $40.6$ & $73.3$ & $57.7$ & $43.5$& $77.4$\\
			PersonLab~\cite{PapandreouZCGTM18} & 1401 &$66.5$ & $86.2$ & $71.9$ & $62.3$ & $73.2$ & $70.7$ & $65.6$ & $77.9$\\
			HrHRNet-W$32$ + AE ~\cite{cheng2019bottom} & $512$ & $67.1$ & $86.2$ & $73.0$ & $-$ & $-$ & $-$ & $61.5$ & $76.1$ \\
			HrHRNet-W$48$ + AE ~\cite{cheng2019bottom} & $640$ & $69.9$ & $87.2$ & $76.1$ & $-$ & $-$ & $-$ & $65.4$ & $76.4$ \\
			\hline 
			Our approach (HRNet-W$32$) & $512$ & $68.0$ & $86.7$ & $74.5$ & $62.1$ & $77.7$ & $73.0$ & $66.2$ & $82.7$\\
			Our approach (HRNet-W$48$) & $640$ & $71.0$ & $88.3$ & $77.4$ & $66.7$ & $78.5$ & $76.0$ & $70.6$ & $84.0$\\
			\hline
			\multicolumn{10}{c}{multi-scale testing}\\
			\hline
			HGG~\cite{JinLXWQOL20} & $512$ &$68.3$ & $86.7$&$75.8$&$-$&$-$&$72.0$ & $-$& $-$\\
			Point-Set Anchors~\cite{weiSLW20} & $640$ & $69.8$ & $88.8$ & $76.3$ & $65.9$ & $76.6$ & $75.6$ & $70.6$ & $83.1$ \\
			HrHRNet-W$32$ + AE ~\cite{cheng2019bottom} & $512$ & $69.9$ & $87.1$ & $76.0$ & $-$ & $-$ & $-$ & $65.3$ & $77.0$ \\
			HrHRNet-W$48$ + AE ~\cite{cheng2019bottom} & $640$ & $72.1$ & $88.4$ & $78.2$ & $-$ & $-$ & $-$ & $67.8$ & $78.3$ \\
			\hline
			Our approach (HRNet-W$32$) & $512$ & $70.7$ & $87.7$ & $77.1$ & $66.2$ & $77.8$ & $75.9$ & $70.5$ & $83.6$\\
			Our approach (HRNet-W$48$) & $640$ & $72.3$ & $88.3$ & $78.6$ & $68.6$ & $78.6$ & $77.7$ & $72.8$ & $84.9$\\
			\hline
		\end{tabular}
		\vspace{-.3cm}
	\end{table*}
	
		\renewcommand{\arraystretch}{1.2}
	\begin{table}[t]
		\centering\setlength{\tabcolsep}{3.53pt}
		\footnotesize
		\caption{GFLOPs and \#parameters 
			of the representative top competitors and our approaches with the backbones: HRNet-W$32$ (DEKR$32$) and HRNet-W$48$ (DEKR$48$).
			AE-HG = associative embedding-Hourglass.
		} 
		\vspace{0.5mm}
		\label{tab:model-size}
		\begin{tabular}{l|ccc|cc}
			\hline
			~ & AE-HG & PersonLab & HrHRNet & DEKR$32$ & DEKR$48$ \\
			\hline
			Input size & $512$ & $1401$ & $640$ & $512$ & $640$ \\
			\hline
			\#param. (M) & $227.8$ & $68.7$ & $63.8$ & $29.6$ & $65.7$  \\
			\hline
			GFLOPs & $206.9$ & $405.5$ & $154.3$ & $45.4$ & $141.5$ \\
			\hline 
		\end{tabular} 
		\vspace{-.2cm}
	\end{table}

\section{Experiments}
	\label{sec:experiments}
	\subsection{Setting}
	\label{sec:coco}
	\noindent\textbf{Dataset.}
	We evaluate the performance on the COCO keypoint detection task~\cite{LinMBHPRDZ14}. 
	The train$2017$ set includes $57K$ images 
	and $150K$ person instances annotated with $17$ keypoints, 
	the val$2017$ set contains $5K$ images,
	and the test-dev$2017$ set consists of $20K$ images. 
	We train the models on the train$2017$ set and report the results on the val$2017$ and test-dev$2017$ sets.
	
\noindent\textbf{Training set construction.}
The training sets consist of keypoint and center heatmaps,
and offset maps.

\vspace{.03cm}
\noindent\emph{Groundtruth keypoint and center heatmaps:} 
	The groundtruth keypoint heatmaps $\mathsf{H}^*$
	for each image 
	contains $K$ maps,
	and each map corresponds to one keypoint type.
	We build them
	as done in~\cite{NewellHD17}:
	assigning a heat value
	using the Gaussian function
	centered at a point
	around each groundtruth keypoint.
	The center heatmap is similarly constructed
	and described in the following.
	
\vspace{.03cm}
\noindent\emph{Groundtruth offset maps:} 	
The groundtruth offset maps $\mathsf{O}^*$
for each image
are constructed
	from all the poses $\{\mathcal{P}_1, \mathcal{P}_2, \cdots, \mathcal{P}_N\}$.
	We use the $n$th pose $\mathcal{P}_n$ as an example and others are the same.
	We compute the center position $\bar{\mathbf{p}}_n = \frac{1}{K}\sum_{k=1}^K\mathbf{p}_{nk}$
	and the offsets
	$\mathcal{T}_n = \{\mathbf{p}_{n1} - \bar{\mathbf{p}}_n,
	\mathbf{p}_{n2} - \bar{\mathbf{p}}_n,
	\cdots,
	\mathbf{p}_{nK} - \bar{\mathbf{p}}_n
	\}$
	as the groundtruth offsets for
	the pixel corresponding to the center position.
	We use an expansion scheme
	to augment the center point
	to the center region:
	$\{\mathbf{m}_n^1, \mathbf{m}_n^2, \cdots, \mathbf{m}_n^M\}$, 
	which are the central positions around 
	the center point $\bar{\mathbf{p}}_n$
	with the radius $4$,
	and accordingly update the offsets.
	The positions not lying in the region
	have no offset values.

	Each central position $\mathbf{m}_n^m$ has 
	a confidence value $c_n^m$
	indicating how confident it is the center
	and computed using the way forming the
	groundtruth center heatmap $\mathbf{C}^*$\footnote{In case that one position belongs to two or more central regions, we choose only one central region whose center is the closest to that position.}.	
	The positions not lying in the region
	have zero heat value.

	\vspace{.1cm}
	\noindent\textbf{Evaluation metric.}
	We follow the standard evaluation metric\footnote{\url{http://cocodataset.org/\#keypoints-eval}} 
	and use OKS-based metrics
	for COCO pose estimation.
	We report average precision 
	and average recall scores
	with different thresholds
	and different object sizes:
	$\operatorname{AP}$,
	$\operatorname{AP}^{50}$,
	$\operatorname{AP}^{75}$,
	$\operatorname{AP}^M$,
	$\operatorname{AP}^L$,
	$\operatorname{AR}$,
	$\operatorname{AR}^{M}$,
	and $\operatorname{AR}^{L}$.

	\vspace{.1cm}
	\noindent\textbf{Training.}
	The data augmentation follows~\cite{NewellHD17}
	and includes
	random rotation ($[\ang{-30}, \ang{30}] $),
	random scale ($[0.75, 1.5]$) and random translation ($[-40, 40]$). 
	We conduct image cropping
	to $512\times512$ for HRNet-W$32$ or $640\times640$ for HRNet-W$48$ with random flipping as training samples.
	
	We use the Adam optimizer~\cite{KingmaB14}.
	The base learning rate is set as $1\mathrm{e}{-3}$,
	and is dropped to $1\mathrm{e}{-4}$ and $1\mathrm{e}{-5}$ 
	at the $90$th and $120$th epochs, respectively.
	The training process is terminated within $140$ epochs.
	
	\vspace{.1cm}
	\noindent\textbf{Testing.} 
	We resize the short side of the images to $512/640$ and keep the aspect ratio between height and width,
	and compute the heatmap and pose positions by averaging the heatmaps and pixel-wise keypoint regressions of the original and flipped images.
	Following \cite{NewellHD17}, we adopt three scales $0.5, 1$ and $2$ in multi-scale testing.
	We average the three heatmaps over three scales
	and collect the regressed results from the three scales as the candidates.

	\renewcommand{\arraystretch}{1.15}
	\begin{table*}[t]
		\caption{Comparisons on the COCO test-dev set. $^*$ means using refinement. AE: Associative Embedding.}

		\centering\setlength{\tabcolsep}{11.25pt}
		\label{table:coco_test}
		\footnotesize
		\begin{tabular}{l|c|lllll|lll}
			\hline
			Method & Input size &$\operatorname{AP}$ & $\operatorname{AP}^{50}$ & $\operatorname{AP}^{75}$ & $\operatorname{AP}^{M}$ & $\operatorname{AP}^{L}$ & $\operatorname{AR}$ & $\operatorname{AR}^{M}$ & $\operatorname{AR}^{L}$\\
			\hline
			\multicolumn{10}{c}{single-scale testing}\\
			\hline
			OpenPose$^*$~\cite{CaoSWS17} & $-$ &$61.8$ & $84.9$&$67.5$&$57.1$&$68.2$&$66.5$ & $-$& $-$\\
			AE~\cite{NewellHD17} & $512$ &$56.6$ & $81.8$ & $61.8$ & $49.8$ & $67.0$ & $-$ & $-$&$-$ \\
			CenterNet-DLA~\cite{ZhouWK19} & $512$ &$57.9$ & $84.7$ & $63.1$ & $52.5$ & $67.4$ & $-$ & $-$& $-$\\
			CenterNet-HG~\cite{ZhouWK19} & $512$ &$63.0$ & $86.8$ & $69.6$ & $58.9$ & $70.4$ & $-$ & $-$& $-$\\
			MDN$_3$~\cite{VarameshATT20} & $-$ & $62.9$ & $85.1$ & $69.4$ & $58.8$ & $71.4$ & $-$ & $-$ & $-$\\
			PifPaf~\cite{KreissBA19} & $-$   
			&$66.7$ & $-$ & $-$ & $62.4$ & $72.9$ & $-$& $-$& $-$\\
			SPM$^*$~\cite{NieZYF19} & $-$ & $66.9$ & $88.5$ & $72.9$ & $62.6$ & $73.1$ & $-$ & $-$& $-$\\
			PersonLab~\cite{PapandreouZCGTM18} & $1401$ &$66.5$ & $88.0$ & $72.6$ & $62.4$ & $72.3$ & $71.0$ & $66.1$&$77.7$\\
			HrHRNet-W$48$ + AE ~\cite{cheng2019bottom} & $640$ & $68.4$ & $88.2$ & $75.1$ & $64.4$ & $74.2$ & $-$ & $-$ & $-$ \\
			\hline 
			Our approach (HRNet-W$32$) & $512$ & $67.3$ & $87.9$ & $74.1$ & $61.5$ & $76.1$ & $72.4$ & $65.4$ & $81.9$\\
			Our approach (HRNet-W$48$) & $640$ & $70.0$ & $89.4$ & $77.3$ & $65.7$ & $76.9$ & $75.4$ & $69.7$ & $83.2$\\
			\hline
			\multicolumn{10}{c}{multi-scale testing}\\
			\hline
			AE~\cite{NewellHD17} & $512$ &$63.0$ & $85.7$ & $68.9$ & $58.0$ & $70.4$ & $-$ & $-$& $-$ \\
			AE$^*$~\cite{NewellHD17} & $512$ &$65.5$ & $86.8$&$72.3$&$60.6$&$72.6$&$70.2$ & $64.6$& $78.1$\\
			DirectPose~\cite{TianCS19} & $800$ & $64.8$ & $87.8$ & $71.1$ & $60.4$ & $71.5$ & $-$ & $-$ & $-$\\
			SimplePose~\cite{LiSW20} & $512$ & $68.1$ & $-$ & $-$ & $66.8$ & $70.5$ & $72.1$ & $-$ & $-$\\
			HGG~\cite{JinLXWQOL20} & $512$ &$67.6$ & $85.1$&$73.7$&$62.7$&$74.6$&$71.3$ & $-$& $-$\\
			PersonLab~\cite{PapandreouZCGTM18} & $1401$ &$68.7$ & $89.0$&$75.4$&$64.1$&$75.5$&$75.4$ & $69.7$& $83.0$\\
			Point-Set Anchors~\cite{weiSLW20} & $640$ &$68.7$ & $89.9$&$76.3$&$64.8$&$75.3$&$74.8$ & $69.6$& $82.1$\\
			HrHRNet-W$48$ + AE~\cite{cheng2019bottom} & $640$ & $70.5$ & $89.3$ & $77.2$ & $66.6$ & $75.8$ & $-$& $-$& $-$ \\
			\hline
			Our approach (HRNet-W$32$) & $512$ & $69.8$ & $89.0$ & $76.6$ & $65.2$ & $76.5$ & $75.1$ & $69.5$ & $82.8$\\
			Our approach (HRNet-W$48$) & $640$ & $71.0$ & $89.2$ & $78.0$ & $67.1$ & $76.9$ & $76.7$ & $71.5$ & $83.9$\\
			\hline
		\end{tabular}
		\vspace{-.2cm}
	\end{table*}
	
	\subsection{Results}
	
	\noindent\textbf{COCO Validation.} Table \ref{table:coco_val} 
	shows the comparisons of our method 
	and other state-of-the-art methods.
	Table \ref{tab:model-size} presents
	the parameter
	and computation complexities 
	for our approach and the representative top competitors, such as
	AE-Hourglass~\cite{NewellHD17}, PersonLab~\cite{PapandreouZCGTM18}, and HrHRNet~\cite{cheng2019bottom}.
	
	Our approach,
	using HRNet-W$32$ as the backbone, 
	achieves $68.0$ AP score. 
	Compared to the methods with similar GFLOPs, 
	CenterNet-DLA~\cite{ZhouWK19} and PersonLab~\cite{PapandreouZCGTM18} (with the input size $601$), 
	our approach achieves over $9.0$ improvement.
	In comparison to 
	CenterNet-HG~\cite{ZhouWK19} whose model size is far larger than HRNet-W$32$,
	our gain is $4.0$.
	Our baseline result $61.9$ (Table~\ref{tab:ablation-study-final})
	is lower than CenterNet-HG
	that adopts post-processing
	to match the predictions to the closest keypoints
 	identified from keypoint heatmaps.
 	This implies that our gain comes from 
 	our methodology.

Our approach benefits from large input size and large model size. 
	Our approach, with HRNet-W$48$ as the backbone and
	the input size $640$, 
	obtains the best performance $71.0$
	and $3.0$ gain over HRNet-W$32$. 
	Compared with state-of-the-art methods, 
	our approach gets $7.0$
	gain over CenterNet-HG, $4.5$ gain over PersonLab (the input size $1401$), 
	$3.6$ gain over PifPaf~\cite{KreissBA19} 
	whose GFLOPs are more than twice as many as ours,
	and $1.1$ gain over HrHRNet-W$48$~\cite{cheng2019bottom}
	that uses higher resolution representations.
	
Following~\cite{NewellHD17,PapandreouZCGTM18}, we report
	the results with multi-scale testing. 
	This brings about $2.7$ gain for HRNet-W$32$, 
	$1.3$ gain for HRNet-W$48$.
	
	\vspace{.1cm}
	\noindent\textbf{COCO test-dev.} 
	The results of our approach and other state-of-the-art methods 
	on the test-dev dataset
	are presented in Table \ref{table:coco_test}. 
	Our approach 
	with HRNet-W$32$ as the backbone
	achieves $67.3$ AP score, 
	and significantly outperforms the methods with the similar model size. Our approach with HRNet-W$48$ as the backbone 
	gets the best performance $70.0$, 
	leading to $3.5$ gain over PersonLab, $3.3$ gain over PifPaf~\cite{KreissBA19}, 
	and $1.6$ gain over HrHRNet~\cite{cheng2019bottom}. 
	
	With multi-scale testing, 
	our approach
	with HRNet-W$32$
	achieves $69.8$, 
	even better than PersonLab with a larger model size. 
	Our approach with HRNet-W$48$ achieves $71.0$ AP score, 
	much better than associative embedding~\cite{NewellHD17}, $2.3$ gain over PersonLab, and $0.5$ gain over HrHRNet~\cite{cheng2019bottom}.

\subsection{Empirical Analysis}
\noindent\textbf{Ablation study.}
	We study the effects of the two components:
	adaptive activation (AA)
	and separate regression (SR).
	We use the backbone HRNet-W$32$ as an example.
	The observations are consistent for HRNet-W$48$.
	
	The ablation study results are presented in Table~\ref{tab:ablation-study-final}.
	We can observe:
	(1) adaptive activation (AA) achieves the gain $3.5$
	over the regression baseline
	($61.9$).
	(2) separate regression (SR)
	further improves the AP score
	by $2.6$.
	(3) separate regression w/o adaptive activation gets $1.7$ AP gain.
	The whole gain is $6.1$.

\begin{table}[t]
		\centering
        \setlength{\tabcolsep}{8.6pt}
        \footnotesize
            \renewcommand{\arraystretch}{1.2}
			\newcommand{\tabincell}[2]{\begin{tabular}{@{}#1@{}}#2\end{tabular}}
			\caption{Ablation study
			in terms of 
			the AP score, 
			and four types of errors.
			Adaptive activation (AA) gets $3.5$ AP gain
			over the baseline.
			Separate regression (SR) further gets $2.6$ AP gain.
			Adaptive activation and separate regression
			mainly reduce the two localization errors, Jitter and Miss, by $4.6$ and $1.5$.
			The results are from COCO validation with the backbone HRNet-W$32$.
			}
 		    \vspace{1mm}
			\begin{tabular}{ c  c | c | c  c | c  c }
			\hline
			AA & SR &  AP & Jitter & Miss  & Inversion & Swap\\
				\hline
				& &  $61.9$ & $16.4$ & $7.6$ & $3.3$ & $1.0$\\ 
				& \checkmark &  $63.6$ & $15.2$ & $7.2$ & $3.3$ & $1.0$\\ 
				\checkmark & &  $65.4$ & $13.5$  & $6.7$  & $3.1$ & $1.1$ \\
				\checkmark & \checkmark  & $68.0$ & $11.8$ & $6.1$ & $3.0$ & $1.1$\\
				\hline 
			\end{tabular}

			\label{tab:ablation-study-final}
		    \vspace{-0.2cm}
	\end{table}

	We further analyze how each component contributes
	to the performance improvement
	by using the coco-analyze tool~\cite{RonchiP17}. 
	Four error types are studied: (i) \textit{Jitter} error: small localization error; (ii) \textit{Miss} error: large localization error; 
	(iii) \textit{Inversion} error: confusion between keypoints within an instance. 
	(iv) \textit{Swap} error: confusion between keypoints of different instances.
	The detailed definitions are in~\cite{RonchiP17}.

Table~\ref{tab:ablation-study-final}
shows the errors of the four types
for four schemes. 
The two components,
adaptive activation (AA) and separate regression (SR),
mainly influence the two localization errors,
\emph{Jitter} and \emph{Miss}.
Adaptive activation reduces the \emph{Jitter} error
and the \emph{Miss} error
by $2.9$ and $0.9$, respectively.
Separate regression further reduces
the two errors 
by $1.7$ and $0.6$.
The other two errors are almost not changed.
This indicates that
the proposed two components indeed improve
the localization quality.

\vspace{.1cm}
\noindent\textbf{Comparison with grouping detected keypoints.}
It is reported in HigherHRNet~\cite{cheng2019bottom} that
associative embedding~\cite{NewellHD17} with HRNet-W$32$
achieves an AP score $64.4$ on COCO validation.
The regression baseline using the same backbone HRNet-W$32$
gets an lower AP score $61.9$
(Table~\ref{tab:ablation-study-final}).
The proposed two components lead to an AP score $68.0$,
higher than associative embedding $+$ HRNet-W$32$.

\vspace{.1cm}
\noindent\textbf{Matching regression
to the closest keypoint detection.}
The CenterNet approach~\cite{ZhouWK19}
performs a post-processing step
to refine the regressed keypoint positions
by absorbing the regressed keypoint
to the closest keypoint
among the keypoints identified
from the keypoint heatmaps.

We tried this absorbing scheme.
The results are presented in Table~\ref{table:absorbresults}.
We can see that 
the absorbing scheme does not improve the performance
in the single-scale testing case.
The reason might be that
the keypoint localization quality of our approach 
is very close to that of keypoint identification
from the heatmap.
In the multi-scale testing case,
the absorbing scheme improves the results.
The reason is that
keypoint position regression
is conducted separately
for each scale
and the absorbing scheme
makes the regression results
benefit from the heatmap
improved from multiple scales.
Our current focus is not on multi-scale testing 
whose practical value is not as high as single-scale testing.
We leave finding a better multi-scale testing scheme as our future work.

	\renewcommand{\arraystretch}{1.3}
	\begin{table}[t]
		\caption{Match regression
		to the closest keypoints
		detected from the keypoint heatmaps. 
		Matching does not improve the single-scale
		(ss) testing performance,
		and helps multi-scale (ms) testing.
		Direct regression may need a proper
		multi-scale testing scheme,
		which leaves as our future work. 
		D-$32$ = DEKR with HRNet-W$32$. 
		D-$48$ = DEKR with HRNet-W$48$. 
		}
		\centering\setlength{\tabcolsep}{5pt}
 		\vspace{1mm}
		\label{table:absorbresults}
		\small
		\begin{tabular}{l|cc|cc}
		    \hline
		   	~ & D-$32$ (ss) & D-$48$ (ss) & D-$32$ (ms) & D-$48$ (ms)\\
			\hline
			COCO Val & 
            $68.0_{-0.0}$ & $71.0_{-0.0}$ & $71.0_{ \textcolor{red}{\uparrow 0.3}}$ & $72.8_{\textcolor{red}{\uparrow 0.5}}$\\
			\hline
			COCO Test & 
			$67.3_{-0.0}$ & $70.1_{\textcolor{red}{\uparrow 0.1}}$ & $70.2_{\textcolor{red}{\uparrow 0.4}}$ & $71.4_{\textcolor{red}{\uparrow 0.4}}$ \\
			\hline
			CrowdPose & 
		    $65.5_{\textcolor{blue}{\downarrow 0.2}}$ & $67.0_{\textcolor{blue}{\downarrow 0.3}}$ & 
			$67.5_{\textcolor{red}{\uparrow 0.5}}$ & $68.3_{\textcolor{red}{\uparrow 0.3}}$\\
			\hline
		\end{tabular}
		\vspace{-0.2cm}
	\end{table}
	
\subsection{CrowdPose}
    \noindent\textbf{Dataset.}
    We evaluate our approach on the CrowdPose \cite{li2018crowdpose} dataset
    that is more challenging and 
    includes many crowded scenes. The train set contains $10K$ images, the val set includes $2K$ images and the test set consists of $20K$ images. We train our models on the CrowdPose train and val sets and report the results on the test set as done in~\cite{cheng2019bottom}.
    
	\vspace{.1cm}
	\noindent\textbf{Evaluation metric.}
	The standard average precision based on OKS which is the same as COCO is adopted as the evaluation metrics. The CrowdPose dataset is split into three crowding levels: easy, medium, hard.  
	We report the following metrics:
	$\operatorname{AP}$,
	$\operatorname{AP}^{50}$,
	$\operatorname{AP}^{75}$,
	as well as
	$\operatorname{AP}^E$, $\operatorname{AP}^M$
	and $\operatorname{AP}^H$
	for easy, medium and hard images.
	
	\vspace{.1cm}
	\noindent\textbf{Training and testing.}
	The train and test methods follow 
	those for COCO except the training epochs.
	We use the Adam optimizer~\cite{KingmaB14}.
	The base learning rate is set as $1\mathrm{e}{-3}$,
	and is dropped to $1\mathrm{e}{-4}$ and $1\mathrm{e}{-5}$ 
	at the $200$th and $260$th epochs, respectively.
	The training process is terminated within $300$ epochs.
	
	\vspace{.1cm}
	\noindent\textbf{Test set results.} 
	The results of our approach and other state-of-the-art methods 
	on the test set
	are showed in Table \ref{table:crowdpose_test}. 
	Our approach 
	with HRNet-W$48$ as the backbone achieves $67.3$ AP
	and is better than HrHRNet-W$48$ ($65.9$)
	that is a keypoint detection and grouping approach 
	with a backbone that is designed for
	improving heatmaps.
	With multi-scale testing, 
	our approach 
	with HRNet-W$48$ achieves 68.0 AP score
	and by a further matching process
	(see Table~\ref{table:absorbresults})
	the performance is improved, leading to 0.7 gain over
	HrHRNet-W$48$~\cite{cheng2019bottom}.

\renewcommand{\arraystretch}{1.2}
\begin{table}[t]
		\caption{Comparisons on the CrowdPose test set.}
		\centering\setlength{\tabcolsep}{1.64pt}
		\label{table:crowdpose_test}
		\footnotesize
		\begin{tabular}{l|c|llllll}
			\hline
			Method & Input size &$\operatorname{AP}$ & $\operatorname{AP}^{50}$ & $\operatorname{AP}^{75}$ & $\operatorname{AP}^{E}$ & $\operatorname{AP}^{M}$ & $\operatorname{AP}^{H}$\\
			\hline
			\multicolumn{8}{c}{single-scale testing}\\
			\hline
			OpenPose~\cite{CaoSWS17} & $-$ &$-$ & $-$&$-$&$62.7$&$48.7$&$32.3$\\
			HrHRNet-W$48$~\cite{cheng2019bottom} & $640$ & $65.9$ & $86.4$ & $70.6$ & $73.3$ & $66.5$ & $57.9$ \\
			\hline 
			Ours (HRNet-W$32$) & $512$ & $65.7$ & $85.7$ & $70.4$ & $73.0$ & $66.4$ & $57.5$\\
			Ours (HRNet-W$48$) & $640$ & $67.3$ & $86.4$ & $72.2$ & $74.6$ & $68.1$ & $58.7$\\
			\hline
			\multicolumn{8}{c}{multi-scale testing}\\
			\hline
			HrHRNet-W$48$~\cite{cheng2019bottom} & $640$ & $67.6$ & $87.4$ & $72.6$ & $75.8$ & $68.1$ & $58.9$ \\
			\hline
			Ours (HRNet-W$32$) & $512$ & $67.0$ & $85.4$ & $72.4$ & $75.5$ & $68.0$ & $56.9$\\
			Ours (HRNet-W$48$) & $640$ & $68.0$ & $85.5$ & $73.4$ & $76.6$ & $68.8$ & $58.4$\\
			\hline
		\end{tabular}
		\vspace{-.2cm}
	\end{table}
	
\section{Conclusions}
The proposed direct regression approach
DEKR 
improves the keypoint localization quality
and 
achieves 
state-of-the-art bottom-up pose estimation results.
The success stems from
that we disentangle the representations
for regressing different keypoints
so that each representation 
focuses on the corresponding keypoint region.
We believe that the idea of regression
by focusing and disentangled keypoint regression
can benefit some other methods,
such as 
CornetNet~\cite{LawD18}
and CenterNet~\cite{DuanBXQHT19}
for object detection.

\end{document}